\documentclass[10pt,journal]{IEEEtran}
\usepackage[font=small, labelsep=space]{caption}
\usepackage{verbatim}
\usepackage{graphicx}
\usepackage{multirow}
\usepackage{amssymb}
\usepackage{amsmath}
\usepackage{amsthm}
\usepackage{amsfonts}
\usepackage{enumitem}
\usepackage{paralist}
\usepackage{stfloats}
\usepackage{url}
\usepackage{epstopdf}
\usepackage{float}
\usepackage{color}
\usepackage{cite}
\usepackage{paralist}
\usepackage[noend]{algpseudocode}
\usepackage{algorithmicx}
\usepackage{algorithm}
\usepackage{mathrsfs}
\usepackage{amsmath}
\usepackage{hyperref}
\hypersetup{hypertex=true,
            linkcolor=red,
            anchorcolor=red,
            citecolor=red}
\usepackage{subfigure}
\usepackage{amsmath}
\allowdisplaybreaks[4]
\let\itemize\compactitem
\let\enditemize\endcompactitem
\let\enumerate\compactenum
\let\endenumerate\endcompactenum

\begin{document}
\title{{DRL-Enabled Trajectory Planing for UAV-Assisted VLC: Optimal Altitude and Reward Design}}
\author{Tian-Tian Lin, Yi Liu, Xiao-Wei~Tang, Yunmei Shi, Yi Huang, Zhongxiang Wei, Qingqing Wu, Yuhan Dong
\thanks{Tian-Tian Lin, Yi Liu, Xiao-Wei Tang, Yunmei Shi, Yi Huang, Zhongxiang Wei ({\{2151128, 2250692, xwtang, ymshi, huangyi718b, z\_wei\}@tongji.edu.cn}) are with the Department of Information and Communication Engineering, Tongji University, Shanghai, China.}
\thanks{Qingqing Wu (qingqingwu@sjtu.edu.cn) is with the Department of Electronic Engineering, Shanghai Jiao Tong University, Shanghai 200240, China.}
\thanks{Yuhan Dong (dongyuhan@sz.tsinghua.edu.cn) is with the Shenzhen International Graduate School, Tsinghua University, Shenzhen 518055, China.}
}
\maketitle
\begin{abstract}
Recently, the integration of unmanned aerial vehicle (UAV) and visible light communication (VLC) technologies has emerged as a promising solution to offer flexible communication and efficient lighting. This letter investigates the three-dimensional trajectory planning in a UAV-assisted VLC system, where a UAV is dispatched to collect data from ground users (GUs). The core objective is to develop a trajectory planning framework that minimizes UAV flight distance, which is equivalent to maximizing the data collection efficiency. This issue is formulated as a challenging mixed-integer non-convex optimization problem. To tackle it, we first derive a closed-form optimal flight altitude under specific VLC channel gain threshold. Subsequently, we optimize the UAV horizontal trajectory by integrating a novel pheromone-driven reward mechanism with the twin delayed deep deterministic policy gradient algorithm, which enables adaptive UAV motion strategy in complex environments. Simulation results validate that the derived optimal altitude effectively reduces the flight distance by up to 35\% compared to baseline methods. Additionally, the proposed reward mechanism significantly shortens the convergence steps by approximately 50\%, demonstrating notable efficiency gains in the context of UAV-assisted VLC data collection.
\end{abstract}

\section{Introduction}
Unmanned aerial vehicles (UAVs) have expanded broad applications in areas like emergency communication and data collection due to their small size and high flexibility \cite{xwtang1, xwtang2}. Most UAV-enabled work has concentrated on radio frequency (RF) technology, which primarily addressed issues on security concerns, path planning, and power allocation \cite{xwtang3}. However, its efficacy is compromised in environments with limited spectrum resources and pronounced electromagnetic interference. By contrast, combining UAVs with visible light communication (VLC) offers a clever way to bypass the RF band. This solution not only eases spectrum congestion but also provides superior resistance to electromagnetic interference.

In recent years,  remarkable advancements have been made in the academic research concerning  UAV-assisted VLC systems. VLC uses light-emitting diodes (LEDs) as transmitters and can provide high data rate communication with strong anti-interference capabilities and low energy consumption. Wang \emph{et al}. modeled the channel between an LED and a UAV and derived closed-form expressions for the probability density and the cumulative distribution functions of the orientation error \cite{swang}. Cang \emph{et al}. studied the deployment and resource management for VLC-enabled UAV networks \cite{ycang}, where UAVs provided terrestrial users with wireless services and illumination simultaneously. Liu \emph{et al}. further studied the joint spatial deployment and resource allocation for VLC-enabled UAV network with co-channel interference, aiming to maximize the energy efficiency subject to the limited battery capacity of UAVs \cite{jliu}. Li \emph{et al}. proposed a new power allocation strategy for VLC-UAV networks, with the objective of optimizing the sum-throughput while considering users' quality of service, illumination requirements, eye safety and transmission power constraints \cite{qli}.

Trajectory/location optimization for UAVs in VLC systems is a critical issue that has been investigated in several studies. In particular, Eltokhey \emph{et al}. conducted UAV's location optimization in UAV-based VLC networks using particle swarm optimization method, aiming to minimize multi-user interference while avoiding complex handover procedures \cite{mweltokhey}. Wang \emph{et al}. investigated the dynamical deployment of VLC-enabled UAVs via using the deep learning method, which jointly optimized UAV deployment, user association, and power efficiency while meeting the illumination and communication requirements of users \cite{ywang}. Maleki \emph{et al}. applied the multi-agent deep deterministic policy gradient method to optimize the three-dimensional (3D) UAV trajectory and resource management, seeking to maximize the total data rate and minimize the total communication power consumption simultaneously \cite{mrmaleki}. Pang \emph{et al}. investigated the UAV-VLC network deployment by jointly optimizing user association, UAV placement and power allocation. This optimization challenge is mathematically formulated as a minimization of energy consumption problem and addressed using the deep reinforcement learning (DRL) approach \cite{lpang}. Li \emph{et al}. introduced a joint optimization approach encompassing UAV trajectory planning, grouping of users, and allocation of power via adopting a hybrid action space DRL algorithm, aiming at maximizing the overall user rate in UAV-enabled VLC systems \cite{lli}.

Although previous research has explored altitude optimization within the realm of 3D trajectory planning, the predominant solutions typically hinge on numerical methods or simplified iteration models,  thus lacking a theoretically derived optimal altitude based on VLC channel gain. In addition, the reward mechanisms designed for DRL-based trajectory planning fail to fully integrate fundamental VLC operational principles, such as the Lambertian radiation model, thus exhibiting slow convergence. Therefore, we seek to address these gaps in this letter. Our primary investigation pertains to 3D trajectory planning for UAV-assisted VLC systems, wherein the UAV is deployed to collect data from several ground users (GUs). This problem is solved via an improved twin delayed deep deterministic policy gradient (TD3) algorithm, which is suitable for high-dimensional continuous action spaces. Our main contributions include the following two aspects. Firstly, we derive an optimal UAV flight altitude via thoroughly analyzing the VLC channel gain model, which provides a theoretical basis for 3D UAV trajectory planning and greatly reduces the UAV's flight distance. Moreover, we design an innovative reward mechanism for the DRL algorithm by evaluating the successful communication criteria for VLC system, which significantly improves the convergence speed of the DRL algorithm.

\section{System Model and Problem Formulation}
\subsection{System Model}
As illustrated in Fig. 1, we consider a UAV-enabled VLC system which comprises one UAV and $I$ GUs, which are assumed to distributed randomly in a given $D\times D\ \text{m}^2$ geographical area. The locations of the $i$-th GU is denoted by $\textbf{w}_i=[x_i, y_i, 0]$. The UAV is deployed to collect data from GUs by equipping with a photodiode capable of transforming the received optical signals into electrical signals. To facilitate the trajectory planning, we discretize the continuous time domain into $N$ time slots with unequal-length duration $\delta_n, n \in \{1,...,N\}$. The 3D coordinate of the UAV at the $n$-th time slot can be denoted as $\textbf{w}_n = [{x_n},{y_n},h]$ with $h$ representing the UAV's flight altitude. According to the Lambertian model in \cite{ywang}, the channel gain between the $i$-th GU and the UAV at the $n$-th time slot can be formulated as
\begin{equation}
    H_{i,n}{=}\begin{cases}
        \frac{(m + 1)A}{2\pi d_{i,n}^2}g(\psi_{i,n})\cos^m(\phi_{i,n})\cos(\psi_{i,n}), &0{\le} \psi_{i,n}{\le}\Psi_c, \\
        0, & \psi_{i,n} > \Psi_c,
    \end{cases}
    \label{E1}
\end{equation}
where $m$ denotes the Lambertian order given by $m = -\ln2/\ln(\cos{\Phi_{1/2}})$ with $\Phi_{1/2}$ denoting the transmitter semi-angle at half power. $d_{i,n} = {\|\textbf{w}_n - \textbf{w}_i\|}$ represents the distance between the UAV and $i$-th GU at the $n$-th time slot. $A$ is the area of the detector in the photodiode.  $\phi_{i,n}$ and $\psi _{i,n}$ represent the irradiance and incidence angles between the UAV and the $i$-th GU at the $n$-th time slot, respectively. $g(\psi_{i,n})$ is the concentrator gain of the optical filter, which can be given by $g({\psi _{i,n}}) = \left\{ \begin{array}{l}
\frac{{{n_r^2}}}{{{{\sin }^2}{\Psi _c}}},0 \le {\psi _{i,n}} \le {\Psi _c},\\
0,{\psi _{i,n}} > {\Psi _c},
\end{array} \right.$ with $n_r$ denoting the refractive index and $\Psi_c$ denoting the half-angle of the field of view at the receiver.
\begin{figure}[htbp!]
\centering
\includegraphics[width=0.49\textwidth]{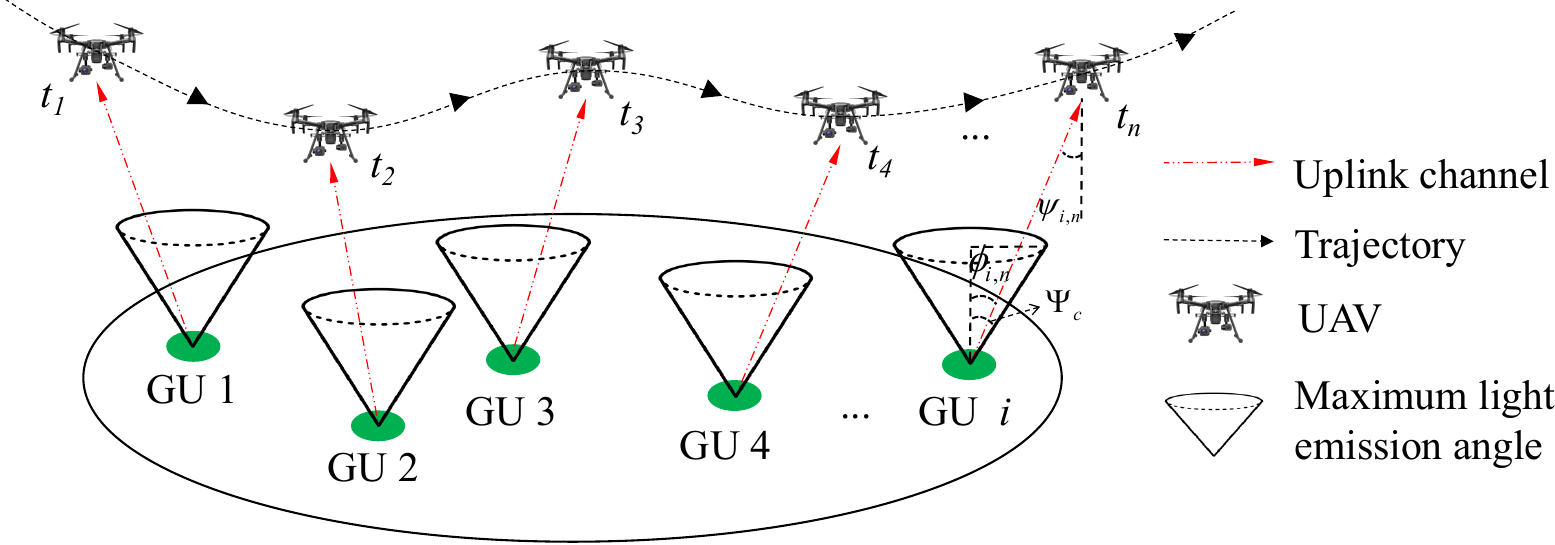}
\caption{UAV-assisted VLC communication system.}
\label{F1}
\end{figure}

Hence, the channel capacity for the VLC link between the UAV and the $i$-th GU, denoted by $C_{i,n}$, can be expressed as
\begin{equation}
    C_{i,n} = \frac{1}{2}\log_2\left( 1 + \frac{e}{2\pi} \left( \frac{\xi P H_{i,n}}{\sigma_w} \right)^2 \right),
    \label{E2}
\end{equation}
where $\xi$ is the illumination response factor of the LED transmitter equipped by GUs, $P$ is the transmission power of GUs, and $\sigma_w$ is the standard deviation of the additive white Gaussian noise.

Let ${b_{i,n}} \in \{0,1\}$ denote the $i$-th GU's coverage metric at the $n$-th time slot, which can be expressed as
\begin{equation}
   b_{i,n} =
   \begin{cases}
   1, & \text{if } H_{i,n} \ge H_{\text{th}}, \\
   0, & \text{otherwise},
   \end{cases}
   \label{E3}
\end{equation}
where $H_{\text{th}}$ denotes the minimum channel gain threshold to ensure that the GU can successfully communicate with the UAV. Furthermore, let $\tilde{b}_{i,n}$ signify whether the $i$-th GU is designated for UAV service during the $n$-th time slot. This can be mathematically formulated as
\begin{equation}
   \tilde b_{i,n} =
   \begin{cases}
   1, & \text{if } b_{i,n} = 1, \text{and }  c_{i,n} = 0, \\
   0, & \text{otherwise},
   \end{cases}
   \label{E4}
\end{equation}
where $c_{i,n} \in \{0,1\} $ indicates whether the $i$-th GU has been served by the UAV and can be given by $c_{i,n} = \min \left\{ \sum_{n' = 0}^n \tilde{b}_{i,n'},\ 1 \right\}$.

\subsection{Problem Formulation}
In this letter, we aim to minimize the UAV's flight distance by optimizing the time slot number $N$ and the UAV's 3D trajectory $\textbf{w}_n$ subject to constraints on maximum serving GU number, task completion, maximum flight speed, flight region, and safe flight altitude, ensuring that the UAV can successively collect all necessary data from GUs without collisions with obstacles. Consequently, the aforementioned problem can be articulated through a mathematical expression as follows.
\begin{subequations}
\begin{align}
(\textbf{P1})~\min_{{\textbf{w}_n},N}~~&\sum_{n = 1}^N d_n, \notag \\
\text{s.t.} \quad &\sum_{i = 1}^I \tilde{b}_{i,n} \le K_{\text{up}},  \forall n, \label{E5a}\\
&\sum_{i = 1}^I c_{i,N} = I, \label{E5b}\\
&0 \le d_n \le d_{\max}, \forall n, \label{E5c}\\
&0 \le x_n, y_n \le D, \quad \forall n, \label{E5d}\\
&h \ge h_{\min}, \label{E5e}
\end{align}
\end{subequations}
where $d_n = \|\textbf{w}_n - \textbf{w}_{n-1}\|$ denotes the UAV's flight distance during the $n$-th time slot. $K_{\text{up}}$ denotes the maximum number of GUs that the UAV can serve at each time slot. Note that if more than $K_{\text{up}}$ GUs is qualified to be served at the $n$-th time slot, the top $K_{\text{up}}$ with the highest channel gain will be chosen to be served. $d_{\max}$ is the maximum distance that the UAV can fly during each time slot, and $h_{\min}$ denotes the minimum flight altitude for the UAV to avoid collision with obstacles. Specifically, (\ref{E5a}) ensures that GUs connect with the UAV at each time slot must not exceed the maximum allowed number $K_{\text{up}}$. (\ref{E5b}) guarantees that all GUs must be served at the end of the flight. (\ref{E5c}) specifies that the UAV's flight distance during each time slot is no larger than the maximum possible distance $d_{\max}$. (\ref{E5d}) require that the UAV flies within the given geographical area. (\ref{E5e}) imposes a constraint on the UAV's safe flight height. It is noteworthy that (P1) is a mixed-integer non-convex problem which is difficult to solve. This challenge is further compounded by the fact that the total number of time slots remains an unknown variable. Moreover, the serving order and achievable channel gain depend on the specific locations of both the UAV and GUs, thus making the above problem intractable to solve by employing traditional optimization methods.

\section{Proposed Solution}
In this letter, we use the TD3 algorithm to solve (P1), which has been demonstrated to be efficient for handling sophisticated control problems in high-dimensional continuous spaces \cite{ywang2}. TD3 employs a delayed policy update mechanism, updating the actor network less frequently than the critic network to enhance training stability. In the following, we first introduce the TD3-based trajectory planning algorithm for (P1). Then, we conduct a theoretical derivation of the optimal flight altitude for the UAV under a given specified channel gain threshold. Lastly, we incorporate a newly designed pheromone-driven reward mechanism into the TD3 algorithm to expedite its convergence speed.

\subsection{TD3-based Trajectory Planning Algorithm}
The TD3 algorithm improves the stability and convergence performance of policy learning by introducing a dual critic network and actor delay update mechanism. It can continuously interact with the environment to update the strategy network and value network, thus achieving a near-optimal UAV motion strategy. The key elements include state $\textbf{s}_n$, action $\textbf{a}_n$, and reward $r_n$, which are detailed as follows.
\begin{enumerate}[1)]
    \item \textbf{State $\textbf{s}_n$}: the state at the $n$-th time slot consists of the GU's coverage state $b_{i,n}$ and serving state $c_{i,n}$, the UAV's location $\textbf{w}_n$, as well as the reward pheromone $\zeta_n$, which is used to guide the exploration direction of UAV and dynamically adjust strategy preferences.
    \item \textbf{Action $\textbf{a}_n$}: the UAV's orientation $\theta_n \in (0, 2\pi]$ and velocity $v_n \in [0,v_{max}]$ are treated as actions.
    \item \textbf{Reward $r_n$}: The task of data collection in this letter is characterized as a sparse-reward problem \cite{ywang2}, whose training difficulty increases exponentially with respect to the number of the GUs when adopting the TD3 algorithm. In detail, the reward can be established as:
        \begin{equation}
            r_n = \begin{cases}
            r_{\tanh}(\zeta_n) + r_{\text{dis}}, & \text{if} \sum\limits_{i = 1}^I c_{i,n} = I, \\
            r_{\tanh}(\zeta_n), & \text{otherwise},
            \end{cases}
            \label{E6}
        \end{equation}
        where ${r_{\tanh }}({\zeta _n}) = [2/(1+\exp(-\zeta_n/(I\cdot\kappa_{\text{cov}})))]-1$ is a shaped reward function of the pheromone $\zeta_n$ with $\kappa_{\text{cov}}$ being a positive constant to denote the UAV's captured pheromone. ${r_{\text{dis}}} = 1/{\sum_{n = 1}^N d_n}$ denotes a reward value that is inversely proportional to the total distance covered by the UAV upon the mission completion.
\end{enumerate}

\subsection{Derivation of the Optimal UAV Flight Altitude}
Based on the VLC channel model presented in Section-II A, an optimal flight altitude can be derived to ensure successful completion of the data collection task while balancing the communication reliability and flight efficiency simultaneously. The detailed derivation process is presented as follows. Under the condition that the UAV and GU are vertical downward and upward, respectively, we have $\cos (\phi_{i,n}) = \cos (\psi _{i,n}) = \frac{h}{d_{i,n}}$. $d_{i,n}$ can be further denoted by $d_{i,n} = \sqrt{h^2 + (d_{i,n}^{xy})^2}$, where $d_{i,n}^{xy}$ denotes the horizontal distance between the UAV and $i$-th GU at the $n$-th time slot. Hence, the channel gain under the condition $0{\le} \psi_{i,n}{\le}\Psi_c$ can be further expressed as
\begin{equation}
{H_{i,n}} = \frac{{(m + 1)Ag({\psi _{i,n}})}}{{2\pi }}\frac{{{h^{m + 1}}}}{{{{({h^2} + {{(d_{i,n}^{xy})}^2})}^{\frac{{m + 3}}{2}}}}}.
    \label{E7}
\end{equation}

Note that the $i$-th GU is successfully served only when the channel capacity $C_{i,n}$ exceeds the minimum required threshold $C_{th}$, i.e., $C_{i,n} \ge C_{th}$, which is also equivalent to $H_{i,n} \ge H_{th}$. In addition, in the context of trajectory planning for UAVs, a greater distance $d_{i,n}^{xy}$ between the UAV and GU implies a shorter flight distance required to complete the task. By letting $H_{i,n}=H_{th}$, we can obtain the basic relationship between $d_{i,n}^{xy}$ and $h$ as $(d_{i,n}^{xy})^2 = {\lambda _{i,n}}{h^{\frac{{2(m + 1)}}{{m + 3}}}} - {h^2}$. For ease of analysis, we define the following $h$-related function.
\begin{equation}
   f(h)  \buildrel \Delta \over =  {\lambda _{i,n}}{h^{\frac{{2(m + 1)}}{{m + 3}}}} - {h^2},
    \label{E8}
\end{equation}
where ${\lambda _{i,n}} = {\left( {\frac{{(m + 1)Ag({\psi _{i,n}})}}{{2\pi {H_{th}}}}} \right)^{\frac{2}{{m + 3}}}}$ is a constant with given $H_{\text{th}}$ under the condition $0{\le} \psi_{i,n}{\le}\Psi_c$. To obtain the optimal altitude $h$ which maximizes $f(h)$, we solve the first and second derivatives of (\ref{E8}) as follows:
\begin{equation}
    f'(h) = \frac{{2(m + 1)}}{{m + 3}}{\lambda _{i,n}}{h^{\frac{{m - 1}}{{m + 3}}}} - 2h,
    \label{E9}
\end{equation}
\begin{equation}
    f''(h) = \frac{{2({m^2} - 1)}}{{{{(m + 3)}^2}}}{\lambda _{i,n}}{h^{ - \frac{4}{{m + 3}}}} - 2.
    \label{E10}
\end{equation}

The zeros of (\ref{E9}) and (\ref{E10}) can be denoted as ${h_0} = {\left( {\frac{{m + 3}}{{{\lambda _{i,n}}(m + 1)}}} \right)^{ - \frac{{m + 3}}{4}}}$ and ${h_{00}} = {\left( {\frac{{{{(m + 3)}^2}}}{{{\lambda _{i,n}}({m^2} - 1)}}} \right)^{ - \frac{{m + 3}}{4}}}$, respectively. Furthermore, since $m \ge 1$, we have  $\frac{{{h_0}}}{{{h_{00}}}} = {\left( {\frac{{m - 1}}{{m + 3}}} \right)^{ - \frac{{m + 3}}{4}}} \ge 1$. We can find that when $h \ge h_{00}$, $f(h)$ monotonically increases within the range of $h_{00} \le h \le h_0$, and monotonically decreases within the range of $h \ge h_0$. We can also find that when $h < h_{00}$, $f(h)$ monotonically decreases with $h$. Therefore, we can obtain the optimal UAV altitude as follows.
\begin{equation}
    h = \left\{ \begin{array}{l}
{h_{\min }},{\rm{if}}\;{h_{\min }} \ge {h_0}\;\text{or}\;f(\;{h_{\min }}) \ge f({h_0}),\\
{h_0},\;\;{\rm{if}}\;f(\;{h_{\min }}) < f({h_0}),
\end{array} \right.
    \label{E11}
\end{equation}

\subsection{Design of the Pheromone-Driven Reward Mechanism}
In the VLC channel gain model given in (\ref{E1}), it can be observed that GU's light signals surpassing the UAV's reception threshold angle (i.e., ${\psi _{i,n}} > {\Psi _c}$) can not be effectively received by the UAV. Given a flight altitude $h$, ${\Psi _c}$ corresponds to a maximum signal reception distance of ${d_{i,n}^{xy,\max}} = h \cdot \tan \left( {{\Psi _c}} \right)$ between the UAV and $i$-th GU. That means the UAV can capture the light signals from the GU when ${d_{i,n}^{xy}} \le {d_{i,n}^{xy,\max}}$. Furthermore, there exists a corresponding threshold ${d_{i,n}^{xy,\text{th}}}$ at the given altitude $h$ according to Section III-B. In traditional TD3 algorithm, the UAV will obtain a reward from the GU only when ${d_{i,n}^{xy}} \le {d_{i,n}^{xy,\text{th}}}$.

However, in instances where ${d_{i,n}^{xy,\text{th}}} < {d_{i,n}^{xy}} \le {d_{i,n}^{xy,\max}}$, the UAV can also receive optical signals from GU but fail to establish a successful communication with GU due to the signal strength below the minimum channel capacity requirements. As a result, the UAV does not receive any rewards in this region. To guide the UAV to continue exploring and flying towards the feasible communication region of GUs, we decide to give it a certain reward when it enters the signal reception region. Therefore, we design the reward pheromone for the UAV as
\begin{equation}
            \zeta_n = \zeta_{n-1} + I_{\mathrm{cov},n} \kappa_{\mathrm{cov}} + \sum\nolimits_{i = 1}^I {\delta (d_{i,n}^{xy})} \kappa_{i,n}^{\text{con}} - \kappa_{\mathrm{dis}} - P_{\mathrm{ob}} ,
            \label{E12}
\end{equation}
where ${\zeta _{n - 1}}$ represents the residual pheromone from the previous $(n-1)$-th time slot. ${I_{{\rm{cov}},n}} = \sum\nolimits_{i = 1}^I {{{\widetilde b}_{i,n}}} $ denotes the number of GUs served by the UAV in the $n$-th time slot.
$\delta(\cdot)$ is a indicator function which can be given by
\begin{equation}
\delta (d_{i,n}^{xy}) = \left\{ \begin{array}{l}
1,\;\text{if}\;d_{i,n}^{xy,\text{th}} < d_{i,n}^{xy} \le d_{i,n}^{xy,\max }\;\text{and}\;{c_{i,n}} = 0,\\
0,\;\text{otherwise},
\end{array} \right.
\label{E13}
\end{equation}
and $\kappa_{i,n}^{\text{con}}$ represents the extra reward pheromone for the UAV entering into the signal reception region. To be specific, $\kappa_{i,n}^{\text{con}}$ is proportional to the residual horizontal distance between the UAV and $i$-th GU within two adjacent time slots, which can be given by $\kappa _{i,n}^{{\rm{con}}} = \rho (d_{i,n}^{xy} - d_{i,n - 1}^{xy})$ with $\rho$ being a positive constant. ${\kappa _{{\rm{dis}}}}$ is a positive constant signifying the fixed loss of pheromone at each time slot. $P_{\text{ob}}$ serves as a penalty if an action results in a violation of the UAV boundary. The extra reward pheromone can effectively decrease the search range of UAV, thereby accelerating the convergence speed of reinforcement learning. The overall algorithm is given in \textbf{Algorithm 1}.

\begin{algorithm}[htb]
\caption{TD3-TDCTM}
\begin{algorithmic}[1]
\Require{Critic networks $\boldsymbol{c}_{\theta_1}, \boldsymbol{c}_{\theta_2}$, actor network $\boldsymbol{\pi}_{\phi}$, target critic networks  $\boldsymbol{c}_{\theta_1'}, \boldsymbol{c}_{\theta_2'}$, target actor network $\boldsymbol{\pi}_{\phi'}$, experience replay buffer $\boldsymbol{R}$.}
\Ensure{The optimal UAV trajectory}.
\State Calculate the optimal flight altitude $h$ according to (\ref{E11}).
\State Let $\theta_1' \leftarrow \theta_1$, $\theta_2' \leftarrow \theta_2$, $\phi' \leftarrow \phi$.
\State \textbf{for} {{episode} = 0 to $N_{\max}$} \textbf{do}
\State  ~~Initialize the state $\boldsymbol{s}_0$ and the pheromone $\zeta_0$.
\State  ~~{\textbf{repeat}}
\State  ~~~~Select an action $\boldsymbol{a}_n = \boldsymbol{\pi}_{\phi}(\boldsymbol{s}_n) + \sigma\varepsilon$, where $\sigma$ is a delay constant and $\varepsilon$ is a Gaussian noise.
\State  ~~~~Update the pheromone $\zeta_n$ according to (\ref{E12}).
\State  ~~~~Observe a reward $r_n$ according to (\ref{E6}).
\State  ~~~~Update the state $\boldsymbol{s}_{n+1}$.
\State  ~~~~Check if the task is completed, i.e., $\sum_{n' = 0}^n {c}_{i,n'} =I$. If ture, update $d_n = 1$ and terminate the episode.
\State  ~~~~Store transition $(\boldsymbol{s}_n, \boldsymbol{a}_n, r_n, \boldsymbol{s}_{n+1}, d_n)$ in $\boldsymbol{R}$.
\State  ~~~~\textbf{if} {$|R| > R_{th}$} \textbf{then}
\State  ~~~~Sample mini-batch of $\boldsymbol{B}$ from $\boldsymbol{R}$.
\State  ~~~~Update $\boldsymbol{c}_{\theta_1}, \boldsymbol{c}_{\theta_2}$, $\boldsymbol{\pi}_{\phi}$, $\boldsymbol{c}_{\theta_1'}, \boldsymbol{c}_{\theta_2'}$, and $\boldsymbol{\pi}_{\phi'}$.
\State  ~~~~\textbf{end}
\State  ~~~~Update $n \leftarrow n + 1$.
\State  ~~\textbf{until} $n = N_{\max}$ or $\sum_{n' = 0}^n {c}_{i,n'} =I$.
\State  \textbf{end}
\end{algorithmic}
\end{algorithm}

\section{Numerical Results}
In the simulation, unless otherwise specified, we set $K_{\text{up}} = 1$, $\Phi_{1/2} = \Psi_c = 60^{\circ}$, $P = 10$ W, $\sigma_w^2 = -128.82$ dBm, $\xi = 0.9$, $d_{\max} = 2$ m, $C_{\text{th}} = 10$, $D = 100$ m, $n_r = 1.5$, $A = 1$c$\text{m}^2$, $h_{\min} = 10$ m, $\varepsilon  \sim \mathcal{N}(0,0.36)$, $\sigma = 0.999$, $|B| = 256$, $R_{th} = 2000$, and $N_{\max} = 8000$. Furthermore, we implement the simulation under the cases of different number of GUs, i.e., $I = 10, 15, 20, 25, 30$.
\begin{figure*}[htbp!]
\centering
\includegraphics[width=0.9\textwidth]{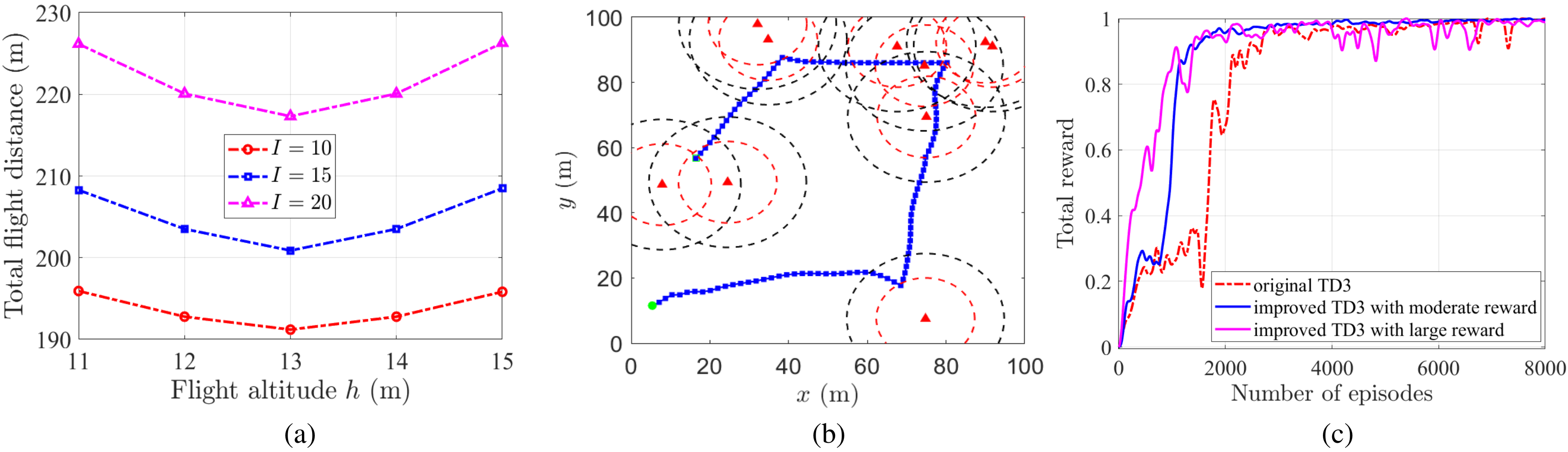}
\caption{Numerical results: (a) relationship between the UAV's flight distance and its altitude, (b) the UAV's horizontal trajectory, and (c) performance of the convergence acceleration.}
\label{fig:2}
\end{figure*}

Fig. \ref{fig:2}(a) illustrates the UAV's total flight distance under different altitudes with 10, 15 and 20 GUs, respectively. It can observed from Fig. \ref{fig:2}(a) that there exists an optimal flight altitude for the UAV to effectively minimize its flight distance, which verifies the theoretical derivation of the optimal flight altitude for the UAV as demonstrated in Section III-B. In addition, we can find that the optimal altitude for the three cases is about $h = 13$ m, which is not affected by the number and distribution of GUs. This is because the optimal altitude is only determined by $m$ and $H_{\text{th}}$, which can be indicated by the closed-form expression given by (\ref{E11}).

Fig. \ref{fig:2}(b) depicts the UAV's horizontal trajectory under the case of $I = 10$, where the red triangle denotes the GU, the red dashed circle signifies the successful communication region, and the black dashed circle indicates the signal reception region. The UAV can receive an additional reward when the UAV enters the circular area between the red and black dashed circles. In the serving process for the GU located at the lower-right corner, it can be observed that the UAV first accelerates its movement towards the red circle when it enters the signal reception region and the UAV promptly proceeds to the next GU upon touching the red circle. This observation fully demonstrates the effectiveness of the designed pheromone-driven reward mechanism.

Fig. \ref{fig:2}(c) illustrates the convergence performance of the original TD3 algorithm and the improved TD3 algorithm with suitable and large reward, respectively. The results demonstrate that the implementation of suitable additional reward can significantly decrease the required time for the algorithm convergence from approximately 4000 time slots to roughly 2000 time slots, thus accelerating convergence by 50\%. Furthermore, it can be observed that large extra reward may cause violent oscillations even when the reward also approaches the final convergence value. Therefore, it is necessary to design a moderate reward pheromone to motivate the UAV to approach the feasible communication region when entering the extra reward region.
\begin{figure}[htbp!]
\centering
\includegraphics[width=0.35\textwidth]{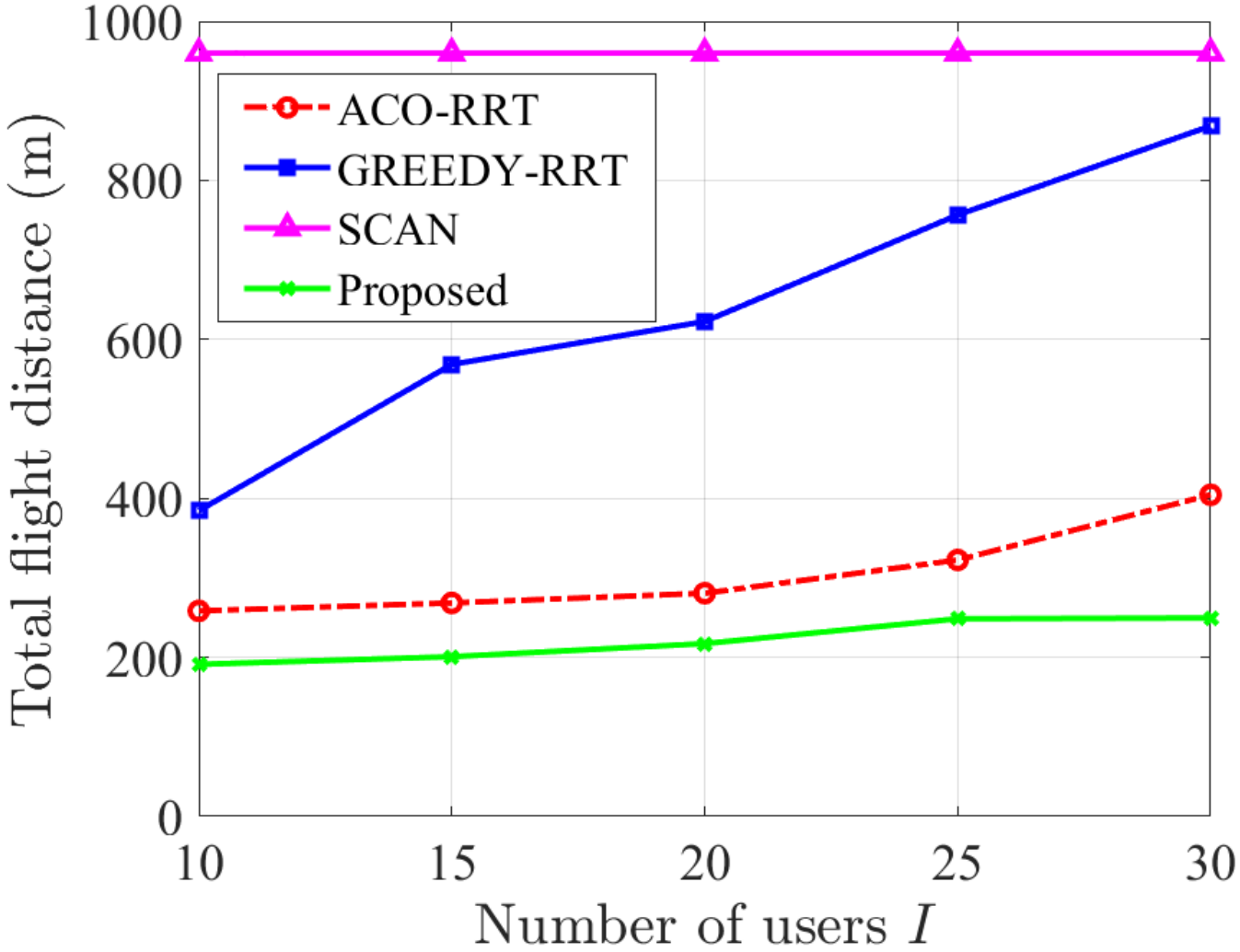}
\caption{Comparison of total flight distance with different algorithms.}
\label{fig:3}
\end{figure}

In this letter, we compare the proposed improved TD3 algorithm with three benchmarks, named SCAN, GREEDY-RRT and ACO-RRT, respectively. Specifically, SCAN algorithm adopts the scanning strategy when designing the UAV's horizontal trajectory. ACO-RRT algorithm adopts the ant colony optimization (ACO) algorithm to select the GUs that should be served and adopts the rapidly-exploring random tree (RRT) algorithm to solve the trajectory planning problem. GREEDY-RRT algorithm is similar to ACO-RRT algorithm, but adopts the greedy algorithm to traverse all the GUs. Note that the UAV flight altitude of three benchmarks is set the same as the proposed scheme. Fig. \ref{fig:3} illustrates the efficiency of the improved TD3 algorithm in Section-III for addressing the UAV-assisted data collection task utilizing VLC technology. Specifically, the total flight distance of the SCAN algorithm stays unchanged under the cases of different numbers of GUs. The UAV's flight distance by adopting the GREEDY-RRT algorithm is nearly linear to the number of GUs. The performance of the ACO-RRT algorithm is superior to that of the SCAN algorithm and the GREEDY-RRT algorithm, and is close to the performance of the proposed scheme. However, compared with ACO-RRT algorithm, the proposed scheme can still achieve  a performance gain of at least 22.3\% in terms of the minimum flight distance.

\section{Conclusions}
In this letter, we proposed an improved TD3 algorithm for UAV trajectory planning in VLC systems, aiming to minimize the UAV's flight distance. By theoretically deriving the optimal UAV altitude based on the VLC channel model, we established a closed-form solution that balanced communication reliability and flight efficiency. Furthermore, we designed a pheromone-driven reward mechanism for the TD3 algorithm to efficiently plan the UAV's horizontal trajectory. Numerical results demonstrate that the proposed solution can reduce the time needed for the algorithm's convergence by approximately 50\% and achieve a performance gain of at
least 22.3\% in terms of the flight distance compared to the state of the art benchmarks.

\vfill

\bibliographystyle{IEEEtran}
\end{document}